\documentclass[letterpaper, 10 pt, conference]{ieeeconf}  

\IEEEoverridecommandlockouts                              

\usepackage{graphics} 
\usepackage{epsfig} 
\usepackage{mathptmx} 
\usepackage{times} 
\usepackage{amsmath} 
\usepackage{amssymb}  
\usepackage[ruled,vlined]{algorithm2e}
\usepackage{multirow}
\usepackage{color}
\usepackage{url}
\usepackage{booktabs}

\usepackage{hyperref}       
\usepackage{graphicx}       
\usepackage{subcaption}     
\usepackage[bottom]{footmisc}

\title{\LARGE \bf
Benchmarking Safe Deep Reinforcement Learning \\in Aquatic Navigation
}

\author{Enrico Marchesini$^{*}$, Davide Corsi$^{*}$ and Alessandro Farinelli
\thanks{*equal contribution}
\thanks{Authors are with the Department of Computer Science, University of Verona, 37135 Verona, Italy. {\tt\small name.surname@univr.it}}
}

\begin{document}

\maketitle
\thispagestyle{empty}
\pagestyle{empty}


\begin{abstract}
We propose a novel benchmark environment for Safe Reinforcement Learning focusing on aquatic navigation. 
Aquatic navigation is an extremely challenging task due to the non-stationary environment and the uncertainties of the robotic platform, hence it is crucial to consider the safety aspect of the problem, by analyzing the behavior of the trained network to avoid dangerous situations (e.g., collisions). To this end, we consider a value-based and policy-gradient Deep Reinforcement Learning (DRL) and we propose a crossover-based strategy that combines gradient-based and gradient-free DRL to improve sample-efficiency. Moreover, we propose a verification strategy based on interval analysis that checks the behavior of the trained models over a set of desired properties. Our results show that the crossover-based training outperforms prior DRL approaches, while our verification allows us to quantify the number of configurations that violate the behaviors that are described by the properties. Crucially, this will serve as a benchmark for future research in this domain of applications.
\end{abstract}
\section{Introduction}
\label{sec:introduction}

Successful applications of Deep Reinforcement Learning techniques in real scenarios are driven by the presence of physically realistic simulation environments \cite{sim_env_gym, sim_env_mujoco, sim_env_safegym}. Along this line, the recent Unity toolkit \cite{sim_env_unity} enables the development of physical simulations to train DRL policies. 

In this paper, we consider the well-known DRL task of robotic navigation \cite{navigation_discretedrl, navigation_drl_1} to introduce a novel aquatic navigation problem, characterized by a physically realistic water surface with dynamic waves. 
In detail, we consider the drones of the EU-funded Horizon 2020 project INTCATCH as a robotic platform. Among the variety of applications for aquatic drones \cite{aquatic_drones_1, aquatic_drones_2}, autonomous water quality monitoring represents an efficient alternative to the more traditional manual sampling \cite{aquatic_drones_intcatch}. To this end, we aim at combining the navigation task and these interesting robotic platforms, to train a robust policy that can autonomously navigate to random targets, avoiding collisions. Our novel aquatic scenario is therefore an important asset to benchmark Safe DRL approaches in a challenging scenario.

A key strength of DRL models is their ability to generalize from training experiences that allow adapting to previously unknown environments \cite{navigation_drl_2, drl_manipulator}. These solutions, however, present several challenges that prevent a wider utilization of DRL. 
    \begin{figure}[t]
        \centering
		\includegraphics[width=0.35\textwidth]{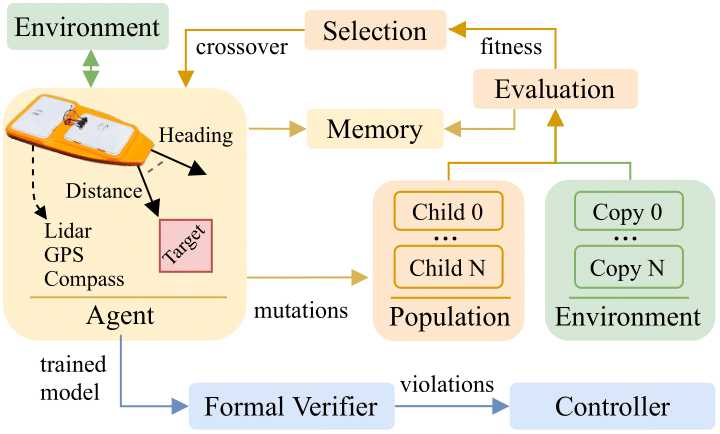}
      	\caption{Overall schematic of our setup.}
    \vspace{-0.6cm}
	\end{figure}
First, DRL has to cope with the environment uncertainties, requiring a huge number of trials to achieve good performance in the training phase. This aspect is particularly important in our scenario where the unpredictable movements of the water significantly increase the complexity of devising robust solutions to the safe navigation problem. We address these issues by extending a recent direction that proposes the combination of gradient-free and gradient-based DRL \cite{mixed_erl, mixed_pderl, mixed_gdrl}, to assimilate the best aspects of both solutions. In more detail, 
we combine a crossover function, typical of Genetic Algorithms (GA), with gradient-based DRL. that improve over prior combined approaches (detailed in Section \ref{sec:background}). The general idea is to incorporate the sampling efficiency of gradient-based DRL while diversifying the collected experiences using the gradient-free population.

Second, robotic tasks usually involve high-cost hardware, hence the behavior of the trained policy must be evaluated to avoid potentially dangerous situations. We address the behavior analysis of these models, by extending prior formal verification \cite{verification_nips} to quantify the occurrences of dangerous situations in the input domain. We also use the state configurations where such undesirable settings occur to discuss how it is possible to exploit this information to design a controller that can avoid such unsafe behaviors.
In more detail, we use interval analysis \cite{verification_interval_analysis} based verification to verify the relations between two or more network outputs in the case of value-based DRL (e.g., in a certain input domain, the network must not select a certain action) or the output values in the case of continuous control. In contrast to prior work that mostly considers such analysis from dependent input intervals, our approach subdivides the analysis into small input intervals that are independent of each other. This allows a straightforward parallelization and the computation of the percentage of undesirable situations, as detailed in Section \ref{sec:method_verifier}. To analyze the behavior of the trained models for aquatic navigation, we introduce a set of properties that describe core behaviors that an autonomous controller should respect in our task. This represents an initial benchmark in aquatic applications, to pave the way for future research about behavior analysis and safety.

Summarizing, we introduce a novel environment for aquatic drones and besides providing a benchmark for the aquatic navigation, our results highlight the following contributions: (i) the crossover-based training improves the performance (i.e., success rate) over a previous combined approach (GDRL \cite{mixed_gdrl}), as well as value-based and policy gradient DRL. Such baselines, in fact, fail at training robust policies in the same number of epochs of the proposed method. (ii) We extend a previous interval analysis tool \cite{verification_nips} to parallelize the evaluation of the properties that describe core behavior for aquatic navigation and compute a \textit{violation} metrics that quantifies the results of this analysis.

\section{Background and Related Work}
\label{sec:background}

    
\subsection{DRL for Robot Navigation and Aquatic Drones}
Achieving autonomy in unstructured environments is a particularly interesting domain of application for DRL as autonomous robotics already achieved tremendous progress in a variety of applications \cite{generic_drones_1}. In this work, we focus on aquatic autonomous navigation \cite{aquatic_drones_1}, where the agent (i.e., an aquatic drone) has to cope with a variety of environment-related problems, such as the inherent instability of the surface (e.g., waves). 
This problem has been partially addressed by standard geometric or model-based techniques \cite{acquatic1, acquatic2}, that can not fully cope with the uncertainties of the operational environment. Hence, we focus on benchmarking aquatic navigation using DRL, which is a well-known learning framework that can adapt to such scenarios.

Recent DRL approaches for robotic navigation, mainly focus on policy-gradient DRL as value-based approaches \cite{algorithm_rainbow} do not scale well in high-dimensional action spaces. Nonetheless, value-based DRL is typically more sample efficient and it has been recently demonstrated that it can provide comparable performance in robotic navigation \cite{navigation_discretedrl}. Hence, to provide a comprehensive benchmark for our aquatic navigation task, we consider both value-based and policy gradient algorithms in our evaluation (Section \ref{sec:method_training}).

\subsection{Combined Deep Reinforcement Learning}
An emergent research trend combines gradient-free  Evolutionary Algorithms and gradient-based DRL, to merge the benefits of both solutions \cite{superl}. In particular, ERL \cite{mixed_erl} considers a DDPG \cite{algorithm_ddpg} agent and a gradient-free training that uses a population of mutated individuals. The DRL agent is trained concurrently from the samples generated by both the approaches and it is periodically injected into the running population. 
Following ERL, several combinations have been proposed. 
Among these, PDERL \cite{mixed_pderl} augments the gradient-free component of ERL, introducing novel operators to improve the genetic representation. In contrast to the actor-critic formalization of these approaches, Genetic DRL (GDRL) \cite{mixed_gdrl} is the first approach that handles both value-based and policy-gradient DRL using a mutation operator. In this work, we augment GDRL with a crossover method to further improve the performance of such approach with both value-based and policy gradient DRL algorithms.

\subsection{Formal Verification for Safe DRL}

Several studies faced the problem of formal analysis for Deep Neural Networks (DNNs) to provide guarantees on their behavior with the analysis of safety properties \cite{safe_survey, prove} on a trained model (i.e., at convergence). A DNN is essentially a function that maps inputs to outputs, through a sequence of layers that apply non-linear transformations (i.e., activation functions). Hence, a standard formulation for the properties adopted by prior work \cite{verification_survey, verification_manipulator}, relates bound on the input values to bounds on the output values of the DNN. The standard formulation of a property is the following:
\begin{equation} \label{equation_1}
\Theta: \mbox{If } x_0\in[a_0, b_0] \land ... \land x_n\in[a_n, b_n] \Rightarrow y\in[c, d]
\end{equation}
\noindent where $x_{k \in [0, n]} \in X$, and $y$ is a generic output.

A promising direction to addresses this problem relies on Moore's interval algebra \cite{verification_interval_analysis}. In particular, ExactReach \cite{interval_method_1} and MaxSense \cite{interval_method_2} represent the first approaches. While the former is an exact method that does not scale to large networks, the latter proposes an approach based on the partitioning of the input domain that suffers from a severe loss of information. Neurify \cite{verification_nips} address these issues and exploits the input space subdivision of MaxSense and the interval analysis propagation of ExactReach, to obtain a scalable and reliable approach. In this work, we extend Neurify by splitting the input space subdivision into sub-intervals that are independent of each other, enabling the parallelization of the method. Our sub-interval structure allows computing a metric that quantifies the overall safety of an agent over the desired properties, as detailed in Section \ref{sec:method_verifier}.
\section{Aquatic Drone Simulator}
\label{sec:method_simulator}

\begin{figure}[t]	
    \vspace{1em}
	\centering
	\begin{subfigure}[t]{0.40\linewidth}
		\centering
		\includegraphics[width=1\linewidth]{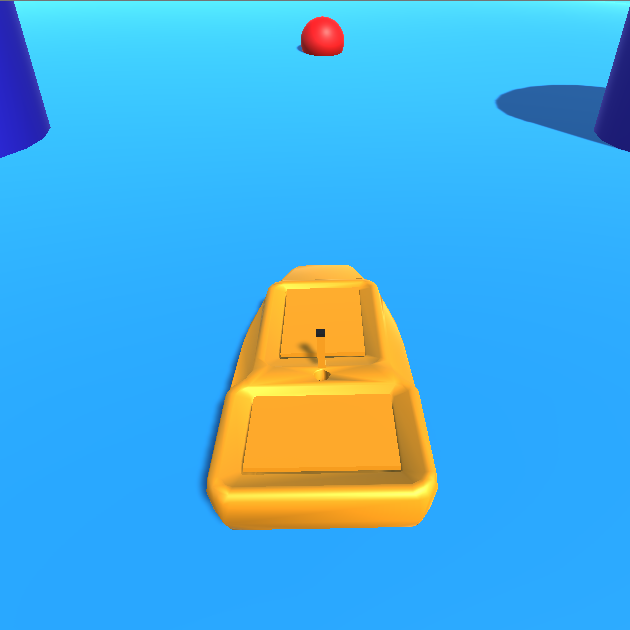}
		\caption{Training simulator}
		\label{fig:aquatic_env_sim}
	\end{subfigure}
	\quad
	\begin{subfigure}[t]{0.40\linewidth}
		\centering
		\includegraphics[width=1\linewidth]{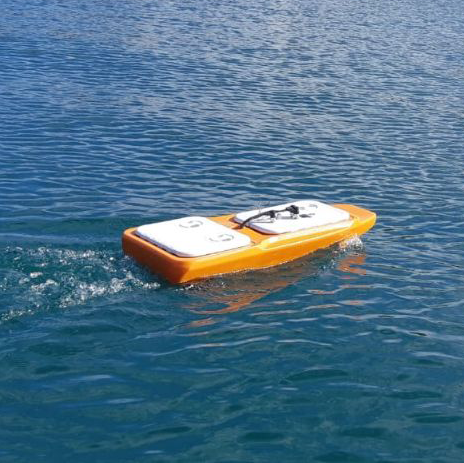}
		\caption{Real aquatic drone}
		\label{fig:aquatic_env_boat}
	\end{subfigure}
	\label{fig:aquatic_env_figures}
	\caption{Our novel Unity water environment and the INTCATCH drone.}
\vspace{-1em}
\end{figure}

In this section, we describe our aquatic navigation task, with an environment characterized by a realistic water surface and dynamic waves.
The Unity physics engine allows us to reproduce the water behavior by triangulating a plane and displacing the generated vertices to simulate the desired wave condition. The plane triangulation can be adjusted to address the trade-off between a more fine-grained simulation of the waves and a higher computational demand imposed on the machine. We considered this, as other particle-based methods such as FleX\footnote{https://developer.nvidia.com/flex} are not suitable to simulate high-dimensional water surfaces.
Trivially, the amount of generated vertices depend on the hardware. Moreover, Unity integrates collision and force algorithms (e.g., gravity, friction, viscosity), that are suitable to simulate the water surface, the aquatic drone, and their interactions. Fig. \ref{fig:aquatic_env_sim} shows an explanatory view of our simulator, where obstacles are represented as blue shapes and the goal as a red sphere. To speed up the training process we consider Unity materials for the rendering pipeline (as they are less computational demanding). However, it is possible to adjust the scene characteristics (e.g., materials, lights, shadows) for a more realistic scene.

The INTCATCH drone that we consider in our work (Fig. \ref{fig:aquatic_env_boat}) is a differential drive platform. It is based on a hull that is equipped with two in-water propellers. The drone can be deployed in shallow water, with a max velocity of $3m/s$. The on-board sensors (e.g., GPS, compass) are used to provide the localization and orientation information, while a lidar is used to collect the distances between the boat and obstacles. 

\subsection{Navigation Description and Network Architecture}
We consider a mapless aquatic navigation problem with obstacles using a dense reward function, and both a discrete and continuous action space (for a more heterogeneous evaluation). The target goals randomly spawn in the scenario and are guaranteed to be obstacle-free. Given the specifications of the drone, the decision-making frequency of the robot is set to $10Hz$, as aquatic navigation does not require high-frequency controls\footnote{Our trained models compute $\approx60$ $actions/s$ on average. Hence, we could increase the control frequency, if required.}. The reward $r_t$ is dense during the travel: $d_{t-1} - d_t$, where $d_{t-1}$, $d_t$ are used to compute the euclidean distance between the robot and the goal at two consecutive time steps. Two sparse values are used in case of reaching the target $r_{reach}=1$ or crashing $r_{fail}=-1$, which terminates an episode (spawning a new target). The discrete action space considers $|y| = 7$ network outputs that are mapped to the power of the two motors, to drive the boat ($y_0, ..., y_2$ for left steering, $y_3$ for forward only movement, $y_4, ..., y_6$ for right steering). 
In the continuous setup, two outputs control each motor velocity ($y_{left}$, $y_{right}$).
Our network takes as input 15 sparse laser scans sampled in $[-90, 90]$ degrees in a fixed angle distribution, and the target position (expressed in polar coordinates). This is a similar setting considered in navigation literature \cite{navigation_discretedrl}. To choose the network size, we performed multiple trials on different seeds and architectures and the outcome led us to use two \textit{ReLU} layers with $64$ neurons and a LSTM layer with 32 hidden units in between the two since we consider a non-stationary environment (this architecture is shared for all the algorithms in Section \ref{sec:training_eval}.

\section{Training}
\label{sec:method_training}
Here we describe our extension to GDRL, which combines gradient-free and gradient-based algorithms with a crossover operator. We performed an initial evaluation to choose the best performing algorithms for the discrete and continuous action spaces. This evaluation led us to choose the value-based Rainbow \cite{algorithm_rainbow} for the discrete scenario, and PPO \cite{algorithm_ppo} for the continuous one. For simplicity of notation, we refer to the resultant algorithms as Population Enhanced Rainbow (PER) and Population Enhanced PPO (PEPPO).

\subsection{Algorithmic Implementation}
An execution of PER and PEPPO proceed as follows: the weights $\theta_{R}$ (or genome) of the agent, are initialized with random values. As in classical training phases, the agent starts to collect experiences interacting with the environment. These samples are stored in a memory buffer, from which are retrieved to train the network. We introduce the gradient-free component periodically by generating a population of networks (or children), each one characterized by different weights. 
Such weights $\theta_{P_i}$ represent the $N$ individuals of the population (with $i \in [1, .. N]$, and $N = 5$ in our experiments) and are generated by applying Gaussian noise to $\theta_{R}$:
\begin{equation}
    \theta_{P_i} = \theta_{R} + mut_p * \mathbb{N}(0, mut_v), \forall i \in [1, .. N]
\end{equation}
\noindent where $mut_v= 0.15$ is the magnitude of the mutation, and $mut_p= 0.2$ is the mutation probability (these values show the best results in our experiments). Hence, we mutate $|\theta_{R}|*mut_p$ weights by a normal distribution with $mut_v$ variance. 

Afterward, a copy of $\theta_{R}$ and the population run over a set of evaluation episodes with hand-designed targets, to find the best individual $\theta_{B}$, based on the fitness. In the case of aquatic navigation, the fitness score is computed as the number of targets that each drone reaches. Ideally, designing a broad variety of targets for this phase is crucial, as we can then assume that $\theta_{B}$ will not return detrimental behaviors to the DRL agent (i.e., it represents an overall better policy). Since the evaluation is independently performed for each $\theta_{P_i}$, we use a separate copy of the environment for each genome in a separate thread, testing them in parallel\footnote{Unity multi-threading makes this phase particularly efficient}. 
During this evaluation, we store a portion of diverse experiences in the memory buffer in the case of the off-policy agent, to incorporate the variety of the population-based component and improve the robustness of the agent. 

Finally, if $\theta_B \in \Theta_P = {\theta_{P_i} | i \in {1, ..., N}}$ (i.e., the DRL agent is not the best policy), $\theta_{R}$ are updated towards the mutated version using our crossover function, and the training phase continues with the new weights. We tried different crossovers, obtaining the best results using a mean crossover similar to Polyak averaging: $\theta_R = \tau \theta_R + (1 - \tau) \theta_{B}$

We tried different values for $\tau$, obtaining the best performance with $\tau = 0.6$. 
In contrast, if $\theta_{B}$ equals the copy of $\theta_{R}$, the training phase of the DRL algorithm continues. 

\subsection{Discussion}
Here we discuss the limitations of prior combined approaches, highlighting the key algorithmic improvements that differentiate our crossover approach:
(i) The gradient-based and gradient-free concurrent training phases proposed in previous work \cite{mixed_erl, mixed_pderl} result in significant overhead. In contrast, PER and PEPPO only perform a periodical evaluation of a population in a multi-thread fashion, reducing the overhead for the training process.
(ii) Prior combination strategies \cite{mixed_erl, mixed_gdrl} do not ensure better performance compared to the DRL agent as they do not prevent detrimental behaviors. 
In contrast, our periodical update of $\theta_{R}$ with the mean crossover is similar to performing a gradient step towards a better policy $\theta_{B}$, limiting detrimental behaviors.

Crucially, our combination mechanism influences the DRL agent policy only if one of its mutated versions performs better in a set of evaluation episodes. Hence, as previously discussed, by hand-designing a sufficient variety of targets for the evaluation, we can obtain a good estimation of the overall policy performance in the population.

    \begin{figure}[b]	
    	\centering
    	\begin{subfigure}[t]{0.45\linewidth}
    		\centering
    		\includegraphics[width=1\linewidth]{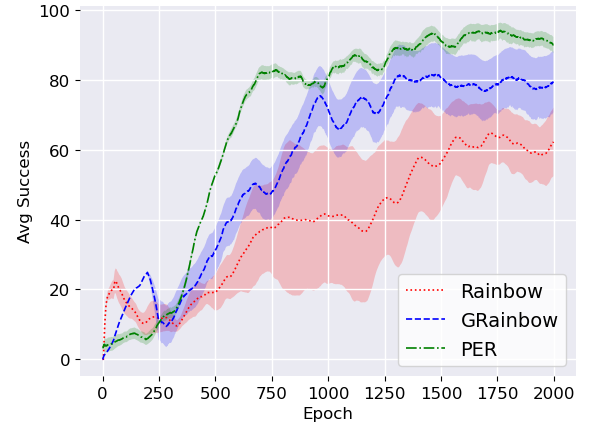}
    		\caption{Discrete Action Space}
    		\label{fig:discrete_success}
    	\end{subfigure}
    	\quad
    	\begin{subfigure}[t]{0.45\linewidth}
    		\centering
    		\includegraphics[width=1\linewidth]{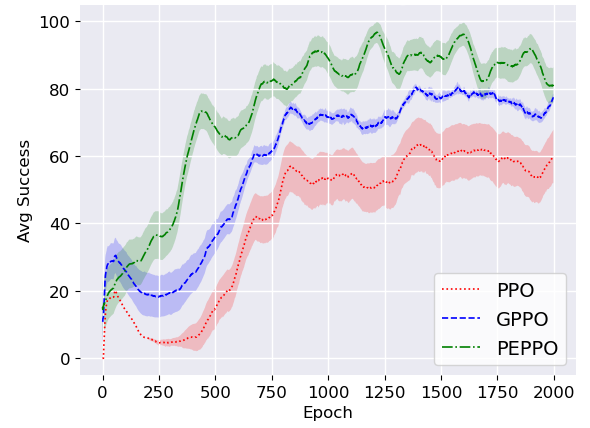}
    		\caption{Continuous Action Space}
    		\label{fig:continuous_success}
    	\end{subfigure}
    	\caption{Average results of PER, PEPPO, and the baselines.}
    	\label{fig:results}
    \vspace{-0.6cm}
    \end{figure}
    
\subsection{Empirical Evaluation}
\label{sec:training_eval}
Our empirical evaluation\footnote{Data are collected on an i7-9700k and a GeForce 2070.} aims at providing a baseline benchmark for our novel DRL aquatic navigation task. We compare our approach with Rainbow and the value-based GDRL (GRainbow \cite{mixed_gdrl}) in the discrete action space scenario, and with PPO and the policy-gradient GDRL (GPPO \cite{mixed_gdrl}) in the continuous one. This will serve as a baseline for future research on aquatic navigation using our novel scenario. For the value-based algorithm we use a prioritized buffer \cite{algorithm_per} of size $50000$ with an $\epsilon$-greedy exploration strategy with $decay=0.995$ and $\epsilon_{min}=0.02$ (this setup shows us the best results). We considered an $\epsilon$-greedy exploration as we already introduce noise in the network weights with our mutations. We refer the interested reader to the original Rainbow \cite{algorithm_rainbow}, and GDRL \cite{mixed_gdrl} for further hyper-parameters details. Our policy-gradient hyper-parameters are the ones presented in the $\epsilon$-clipped version of the original PPO \cite{algorithm_ppo}. 
To collect comparable results when evaluating different algorithms, the random seed is shared across a single run of every algorithm (as there may exist a sequence of targets that favor a run), while it varies in different runs. As a consequence, a certain run of every algorithm executes the same sequence of targets and initializes the networks with the same weights. The following graphs report the mean and standard deviation over ten statistically independent runs (i.e., with different seeds), given the recent work on the importance of the statistical significance of the data collection process \cite{eval_seed}.

For our evaluation we consider the \textit{success rate} metric, that was previously used in robotic navigation \cite{navigation_discretedrl}. This measures how many successful trajectories are performed. Data are smoothed over one hundred epochs for both the discrete (Fig. \ref{fig:discrete_success}) and continuous (Fig. \ref{fig:continuous_success}) action space evaluation.

\begin{figure}[t]	
    	\centering
    	\vspace{1em}
    	\includegraphics[width=0.7\linewidth]{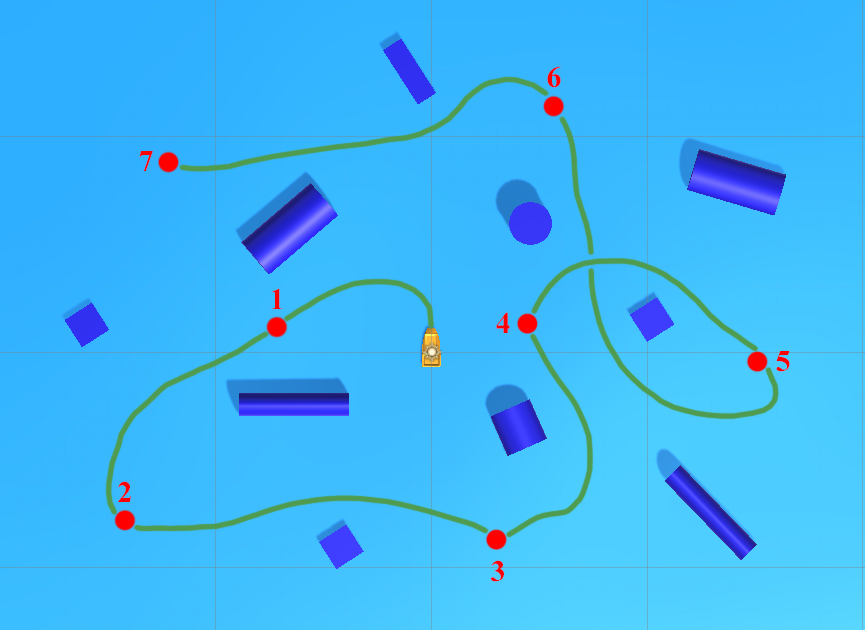}
    	\caption{Explanatory path in a previously unseen scenario.}
    	\label{fig:navigation_eval}
    \vspace{-0.6cm}
    \end{figure}

In detail, the PER reaches over $\approx$95\% successes around epoch 1500, while models trained with the baselines, i.e., Rainbow and GRainbow, do not go beyond a success rate of $\approx$70\% and $\approx$90\%, respectively.
The policy-gradient PEPPO achieves similar results, reaching over $\approx$95\% successes around epoch 1200, while models trained with the baselines, i.e., PPO and GPPO, do not go beyond a success rate of 60\% and 80\%, respectively. Crucially, only models trained with PER and PEPPO are able to converge to a policy that correctly navigates in the environment (i.e., with a success rate over 95\%), considering our limit of 2000 training epochs for this task (increasing the number of epochs leads to higher training times and negligible improvements). Furthermore, the resulting model generalized: velocity, starting, and target position; while the lidar allows the agent to navigate in previously unknown scenarios with different obstacles. Fig. \ref{fig:navigation_eval} shows an explanatory trajectory performed by a trained PER policy in our evaluation environment (considering the same sequence of targets, a converged model with PER and PEPPO return comparable trajectories).

\begin{figure}[b]	
	\centering
	\begin{subfigure}[t]{0.45\linewidth}
		\centering
		\includegraphics[width=1\linewidth]{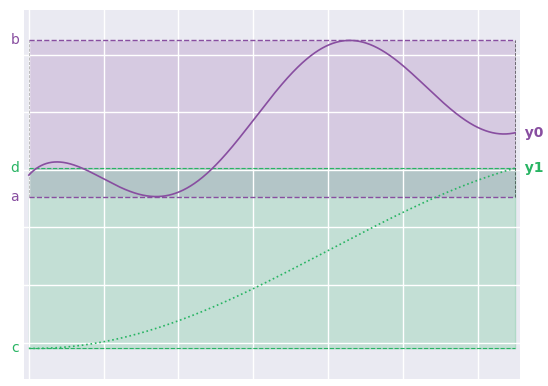}
		\caption{One output function with one subdivision.}
		\label{fig:formal_verifier:a}
	\end{subfigure}
	\quad
	\begin{subfigure}[t]{0.45\linewidth}
		\centering
		\includegraphics[width=1\linewidth]{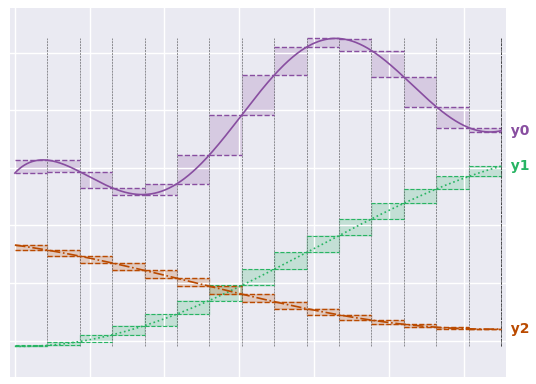}
		\caption{Multiple output curves with a subdivided input area.}
		\label{fig:formal_verifier:b}
	\end{subfigure}
	\caption{Explanatory output analysis.}
	\label{fig:formal_verifier}
    
\end{figure}

\section{Formal Verification}
\label{sec:method_verifier}
In this section, we address the problem of the formal verification of safety properties in our aquatic navigation task. As introduced in Section \ref{sec:background} we consider approaches based on Moore's interval algebra \cite{verification_interval_analysis} to compute strict bounds for the values that each output of the DNN assumes. In detail, these methods propagate a subset of the input domain layer by layer and apply non-linear transformations (when required), to compute the corresponding interval for each output node of the network. Hence, they verify if it lies in the desired range. We refer to the input and output intervals as \textit{input area} and \textit{output bound}, respectively. 
Neurify uses an \textit{iterative refinement} process that subdivides the input area to search input configurations that violate the specified property. Afterward, they perform the union of the output bounds resulting in a more accurate estimation for the original area, addressing the overestimation problem \cite{verification_nips}. In contrast, we directly verify the property on the individual sub-area bounds to make this process completely independent and parallelizable. Hence, we address the overestimation problem on the individual sub-areas. 

\subsection{Deep Reinforcement Learning Navigation Properties}
In a value-based DRL context, the agent typically selects the action to perform that corresponds to the node with the highest value (in the case of value-based DRL).
For this reason, in contrast to the previous properties formalized as Eq. \ref{equation_1} in Section \ref{sec:background}, we propose a slightly different formulation that is explicitly designed for value-based DRL:
\begin{equation} \label{eq:prove_property}
\Theta: \mbox{If } x_0\in[a_0, b_0] \land ... \land x_n\in[a_n, b_n] \Rightarrow y_j > y_i
\end{equation} 

In general, we refer to these properties as \textit{decision properties} as they are used to ensure that a given action (e.g., $y_j$) does not violate the desired behavior. To verify if an interval $y'= [a, b]$ is greater than another one $y''=[c, d]$ , we rely on the interval algebra. In particular, on the equation:
\begin{equation} \label{eq:prop_moore} 
b < c  \Rightarrow y' < y''
\end{equation}

Besides the property formulation for value-based DRL, Fig. \ref{fig:formal_verifier:a} shows the contribution of our extension to Neurify in analyzing a property. Note that we show a scenario where the overestimation problem is perfectly solved (i.e., the output bound [a, b] matches the min and the max of the output function). The figure shows the two different representations of the verification problem, where $y_0$ and $y_1$ represent the output functions generated by two nodes of a generic network. Our idea is to perform the estimation of the output curve for each node, by performing the verification process directly on the sub-intervals. Fig. \ref{fig:formal_verifier:b} summarizes our approach: by subdividing the initial area and calculating the corresponding output bounds for each interval, we obtain a precise estimation of the output curves. By applying Eq. \ref{eq:prop_moore} on each sub-area, we can obtain three possible results from the analysis of a safe decision property: (i) the property holds, (ii) the property is violated, or (iii) we can not assert anything on the property in this area. In the third case, we iterate the partition of the interested area to increase the accuracy of the estimation. Moreover, our approach can compute the portion of the starting area (and eventually returns it as a counterexample) that causes the violation of a specific property. By normalizing this value, we obtain a novel informative metrics, the \textit{violation}. This value represents an upper bound for the real probability that the agent performs undesired behaviors with respect to the property.

\subsection{Decision Properties}
The design of a set of properties for our aquatic navigation scenario that presents a variable set of obstacles is a challenging problem. Given the dynamic and non-stationary nature of such environment, it is not possible to formally guarantee the safety of the drone in any possible situation. For this reason, we focus on ensuring that the agent makes rational decisions (i.e., it selects the best action given available information) to provide an initial benchmark for the safety of this task. Hence, we selected three properties, that represent a possible safe behavior of our agent in relation to possible obstacles:

\textbf{$\Theta_{0}$:} If there is an obstacle near to the left, whatever the target is, go straight or turn right.

\textbf{$\Theta_{1}$:} If there is an obstacle near to the right, whatever the target is, go straight or turn left.

\textbf{$\Theta_{2}$:} If there are obstacles near both to the left and to the right, whatever the target is, go straight. 

To formally rewrite these properties, it is necessary to further detail the structure of the input and the output layers of our DNN, and formally define the concept of \textit{"near"} for an obstacle. In both the value-based and policy-gradient setup, the input layer contains 17 inputs: (i) the lidar collect 15 values normalized $\in [0,1]$, (ii) the heading of the target with respect to the robot heading ($\in [-1, 1]$), and (iii) the distance of the target from the robot ($\in [0, 1]$). The output nodes correspond to specific actions in the value-based case: from $y_0$ to $y_6$ each node represents the action \textit{Strong Right, Right, Weak Right, None, Weak Left, Left, and Strong Left}. While for the policy-gradient scenario we have two outputs $y_{left}$ for the left motor and $y_{right}$ for the right one (for simplicity we only consider forward movements, hence $y_{left}$ and $y_{right}$ can assume values $\in [0, 1]$

Moreover, we measured that an obstacle is close to the agent if its distance is less than $0.35$ on the front, and less than $0.24$ on the two sides (normalized in the corresponding input domain), considering the max velocity of the agent (hence, these values allow us to turn in the opposite direction to a fixed obstacle without colliding). Given this, we obtain the following domain for our properties: 
\begin{center}
$ I_{0}: \mbox{If } x_0, ..., x_5 \in [0.00, 0.24] \land x_5, ..., x_{14}, x_{15}, x_{16} \in \mathbb{D^*} $ \\
$ I_{1}: \mbox{If } x_{10}, ..., x_{14} \in [0.00, 0.24] \land x_0, ..., x_{10}, x_{15}, x_{16} \in \mathbb{D^*} $ \\
$ I_{2}: \mbox{If } x_0, ..., x_5, x_{10}, x_{14} \in[0.00, 0.24] \land x_6, ..., x_6, x_{15}, x_{16} \in \mathbb{D^*} $ 
\end{center}
\noindent where $\mathbb{D^*}$ is the complete domain of the corresponding node where the distance is not \textit{near}. Finally, we formalize the previous decision properties for PER and PEPPO as follow:
\begin{center}
$ \Theta_{0, PER}: \mbox{If } I_0 \Rightarrow [y_4, y_5, y_6] < [y_0, y_1, y_2, y_3] $ \\
$ \Theta_{1, PER}: \mbox{If } I_1 \Rightarrow [y_0, y_1, y_2] < [y_3, y_4, y_5, y_6] $ \\
$ \Theta_{2, PER}: \mbox{If } I_2 \Rightarrow [y_0, y_1, y_5, y_6] < [y_2, y_3, y_4]$ \\
$ \Theta_{0, PEPPO}: \mbox{If } I_0 \Rightarrow y_{left} - y_{right} > k $  \\
$ \Theta_{1, PEPPO}: \mbox{If } I_1 \Rightarrow y_{right} - y_{left} > k $ \\
$ \Theta_{2, PEPPO}: \mbox{If } I_2 \Rightarrow |y_{left} - y_{right}| < k $ \\
\end{center}
\noindent where $k$ is a constant value for the minimum motors power difference that allows a rotation that avoids a collision. 

\subsection{Results}

\begin{table}[t]
\centering
\vspace{1em}
\begin{tabular}{lrrr}
        \toprule
        \textbf{} &  \textbf{Seed 1} & \textbf{Seed 2} & \textbf{Seed 3} \\ 
        \midrule
        $\Theta_{0, PER}$    &   9.3 $\pm$ 2.4    &  5.3 $\pm$ 3.1    & 0.0 $\pm$ 0.0 \\
        $\Theta_{1, PER}$    &   3.0 $\pm$ 2.3    &  4.1 $\pm$ 2.2    & 0.0 $\pm$ 0.0 \\
        $\Theta_{2, PER}$    &   7.1 $\pm$ 1.4    &  6.9 $\pm$ 2.7    & 3.4 $\pm$ 0.2 \\
        \midrule
        $\Theta_{0, PEPPO}$    &   1.3 $\pm$ 0.3 &    1.3 $\pm$ 0.7    & 1.3 $\pm$ 0.2 \\
        $\Theta_{1, PEPPO}$    &   0.2 $\pm$ 0.2 &    0.0 $\pm$ 0.0    & 0.7 $\pm$ 0.5 \\
        $\Theta_{2, PEPPO}$    &   3.6 $\pm$ 1.4 &    6.3 $\pm$ 2.6    & 3.1 $\pm$ 0.1  \\      
    \bottomrule
    \end{tabular}
    \caption{For each property we show the mean and the variance of the violation metric (\%) of the best 5 models for the best 3 seeds (considering the success rate).}
\label{tab:verifier_resutls}
\vspace{-1em}
\end{table}

Table \ref{tab:verifier_resutls} shows the results for each property, considering the average violation of the best ten models of the best performing three seeds used in our training. We focus the best performing algorithms (i.e., PER and PEPPO) to show that converged models that achieve similar return, could suffer from input corner cases in a very different way (e.g., models trained with seed 1 present a violation of $\approx$9.3 on seed 1 for $\Theta_{0, PER}$, while models with seed 3 have a violation of 0 on the same property). Moreover, for each algorithm, we consider the 5 models that achieved the highest success rate during the training phase. Results show that the violation and the success rate are not necessarily related; despite we tested the safety on the best models, we found in some cases a high violation rate. For this reason, our safety analysis is a necessary step to evaluate a policy before its deployment in a real-world environment.
Finally, due to the low violation rate of these models, it is possible to design a simple controller to guarantee the correct behavior of the network. To illustrate this, we describe the process to decide whether the controller can be designed for our navigation task. As described in Section \ref{sec:method_simulator}, the decision-making frequency of the robot is set to $10Hz$ and, with the violation presented in Table \ref{tab:verifier_resutls}, a complete search through the array of the sub-areas that cause a violation always requires less than 0.09s. This means that, with our hardware setup, we can verify if the input state leads to a violation at each iteration, without lags in the robot operations. Consequently, we can avoid all the decisions, derived from input configurations, that lead to the violation of our desired safety properties. 

\section{Conclusion}
\label{sec:conclusion}
We presented a novel aquatic environment characterized by a physically realistic water surface, which we make available as an attachment to the paper along with an explanatory video.
In addition to provide a baseline for a novel DRL aquatic navigation task, we extend a prior DRL algorithm that combines gradient-free and gradient-based approaches with a crossover operator to improve sample-efficiency and performance. We also extend prior formal verification to enable a straightforward parallelization and the computation of a violation metric. We applied these techniques to the challenging problem of aquatic navigation in presence of obstacles, obtaining a safe and robust policy. Finally, this work paves the way for several interesting research directions, which include the possibility to integrate formal verification with our learning approach, to generate safe by construction models.


\bibliographystyle{IEEEtran}
\bibliography{root.bib}


\end{document}